# Points for Energy Renovation (PointER): A LiDAR-Derived Point Cloud Dataset of One Million English Buildings Linked to Energy Characteristics

## Authors

Sebastian Krapf[1], Kevin Mayer[2], Martin Fischer[2]

**Affiliations**

1. Institute of Automotive Technology, Department of Mechanical Engineering, TUM School of Engineering and Design, Technical University of Munich, Boltzmannstr. 15, 85748, Garching b. München, Germany

2. Department of Civil and Environmental Engineering, Stanford University, 473 Via Ortega, 94305, Stanford, USA

corresponding author(s): Sebastian Krapf (sebastian.krapf@tum.de), Kevin Mayer (kdmayer@stanford.edu)

## Abstract

Rapid renovation of Europe's inefficient buildings is required to reduce climate change. However, analyzing and evaluating buildings at scale is challenging because every building is unique. In current practice, the energy performance of buildings is assessed during on-site visits, which are slow, costly, and local. This paper presents a building point cloud dataset that promotes a data-driven, large-scale understanding of the 3D representation of buildings and their energy characteristics. We generate building point clouds by intersecting building footprints with geo-referenced LiDAR data and link them with attributes from UK's energy performance database via the Unique Property Reference Number (UPRN). To achieve a representative sample, we select one million buildings from a range of rural and urban regions across England, of which half a million are linked to energy characteristics. Building point clouds in new regions can be generated with the open source code published alongside the paper. The dataset enables novel research in building energy modeling and can be easily expanded to other research fields by adding building features via the UPRN or geo-location.

## Background & Summary

Buildings are responsible for 40% of the European Union's energy consumption and 36% of its greenhouse gas emissions[1]. In the residential sector, space heating, cooling, and hot water supply constitute up to 80% of citizens' energy consumption. At the same time, more than 75% of EU buildings are inefficient[2], and the share of the building stock that undergoes major renovation is low, ranging from less than 0.4% to 1.2% across EU member states[1]. Therefore, renovating buildings is a key initiative of the European Green Deal[3]. To this end, the European Commission published the "A Renovation Wave for Europe" strategy that aims to tackle inefficient buildings and to double annual energy renovation rates[4].

In practice, Energy Performance Certificates (EPC) are one of the central instruments to contribute to these goals by providing transparency of the existing building stock's energy efficiency. It is mandatory to issue an EPC for buildings up for sale or rent and to display them in advertisements[5]. In Europe, the United Kingdom has the largest number of registered EPCs – more than 20 Million[5] – with around 60% of coverage in England and Wales in 2022[6]. EPCs are created for existing buildings by accredited energy assessors during on-site visits[7], which makes the process slow, costly, and local. To achieve the aggressive renovation targets, methods are needed that generate relevant insights to buildings based on widely available data sources.

To this end, there is increasing research interest, in particular in the field of urban building energy modeling (UBEM)[8–10]. UBEM refers to the bottom-up building energy modeling and analysis on a city-scale. One aim of UBEM is to identify inefficient buildings automatically on large scale with continuous coverage. However, Ali et al. point out that necessary input data is often unavailable for an entire city or a district[9]. Moreover, the prevalent physics-based or engineering approach in UBEM is currently inadequate for large-scale building-level analysis because it requires highly detailed input data[9].

As an alternative to physics-based modeling, some researchers propose data-driven approaches which aim at determining energy characteristics with fewer input data[11–16]. Publications mention different compositions of input features such as occupancy, year of construction, insulation of building envelope, or surface-to-volume ratio. However, these features are usually only available for a specific region or a subset of buildings in an area of interest.

Consequently, researchers turn to extracting relevant features, or proxies that indirectly indicate features, from remotely sensed data, which is available on a larger scale[16–21]. For example, building footprints[16] and roof shapes[18] can be gathered from aerial images. Street view images include information about the window-to-wall ratio or the number of floors[19,21]. Airborne LiDAR contains information about a building's height[15] and can be further processed into Digital Surface Models (DSM)[19,20] or 3D models[22] to reflect a building's envelope and volume. Implicitly, LiDAR also contains information that is relevant for building energy modeling. For example, Tooke et al. predicted building age based on LiDAR[17], and Castagno and Atkins improved roof shape classification with LiDAR[18].

In summary, combining widely available remotely sensed data sources with data-driven algorithms to estimate energy efficiency can provide building insights fast and at scale[21]. LiDAR is a promising data source as it includes features linked to building height, envelope geometry, roof superstructures, and surface-to-volume ratio. In addition, age, building type, or architectural style can be inferred from a building's point cloud representation.

To support and accelerate research in this field, this paper provides a large-scale dataset of building point clouds coupled with energy characteristics. The major contributions are two-fold:

1) The PointER dataset comprises more than 1 million building point clouds and covers 16 diverse authority districts in England. When available, building point clouds are linked to data from UK's energy performance database, leading to more than half a million complete sets of point clouds with energy characteristics. The dataset can be downloaded from https://doi.org/10.14459/2023mp1713501
2) Open source code with a detailed documentation to replicate the building point cloud generation process enables follow-up studies. The process is region-agnostic and can be applied anywhere where LiDAR data and building footprints are available. The code is available at https://github.com/kdmayer/PointER

## Methods

In this chapter, we describe how to acquire and how to store the required base data (Table 1). We also present how we select a representative set of buildings for our dataset and how we generate the building point clouds and link them with energy attributes. Figure 1 gives a high-level overview of the conducted steps to create the PointER dataset.

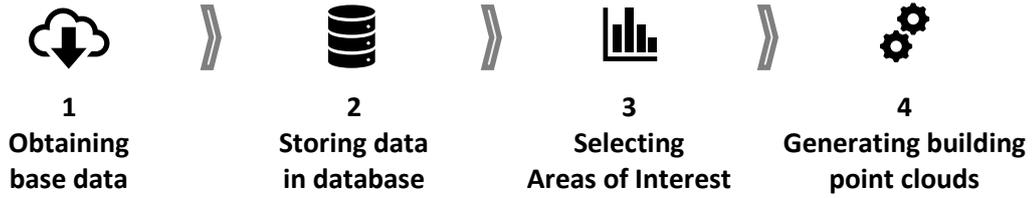

*Figure 1: Data pipeline overview: generating a building point cloud dataset in four steps*

**Obtaining base data**

Multiple open datasets in the UK enable the creation of the PointER building point cloud dataset, including LiDAR data and Europe's largest collection of EPCs. In this study, we focus on England, as not all of the required data is available for the entire UK. To produce the dataset, we obtain seven base datasets which are summarized in Table 1. All datasets are provided under the Open Government License (OGL) except for the building footprints dataset.

The first and major component is UK's National LiDAR Programme Point Cloud data, collected between 2017 and 2021. The LiDAR data covers all of England with a resolution of 1 m and is available online[23]. The point data is categorized into the ASPRS Standard LiDAR Point Classes[24], such as building, ground, low vegetation, water, etc. Hence, building point clouds could simply be extracted from the LiDAR data based on their classification. However, to match energy characteristics on the property level, we require a distinction between neighboring properties. In particular, the boundaries of properties cannot be extracted from LiDAR points alone in the common case of townhouses. Therefore, our alternative approach is to crop points using building footprint data.

*Table 1: Overview of base data sources*

|   | Acronym | Dataset name | Years | License | Ref. |
|---|---------|--------------|-------|---------|------|
| 1 | Point clouds | National LIDAR Programme Point Cloud | 2017 – 2021 | OGL | [23] |
| 2 | Verisk building footprints | UKBuildings edition 13 online version | 2021 | Personal License | [25] |
| 3 | EPC | Energy Performance of Buildings Data: England and Wales | 2008 – 2022 | OGL* | [26] |
| 4 | UPRN | Ordnance Survey Open Unique Property Reference Number | 2022 | OGL | [27] |
| 5 | LAD | Local Authority Districts (Dec 2021) GB BFC | 2021 | OGL | [28] |
| 6 | RUC | 2011 Rural Urban Classification lookup tables for all geographies | 2011 | OGL | [29] |
| 7 | OA | Output Areas (Dec 2011) Boundaries EW (BFC) | 2011 | OGL | [30] |

* except for address information

To this end, we require accurate building footprint data. We chose the UKBuildings edition 13 online version dataset by Verisk[25] because it has the most accurate data based on our visual analysis of aerial images. Figure 2 displays UKBuildings' polygons as well as building outlines from OpenStreetMaps (OSM)[31] and Ordnance Survey's OS OpenMap - Local[32]. As can be seen from the examples, a common downside of OSM is the missing footprint information. Likewise, the OS dataset is not suitable because it does not contain exact property lines between the buildings. While Verisk's footprint data has a higher quality, the drawback is that we cannot publish the footprint data due to the private license.

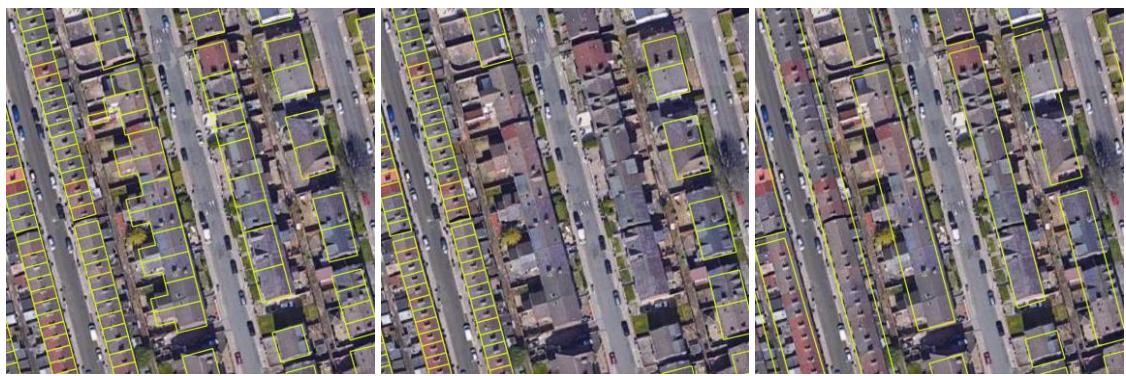

*Figure 2: Building outline polygons of three available datasets visualized on aerial images. UKBuildings[25] (left), OSM[31] (center), and OS OpenMap[32] (right) on Google Maps aerial image.*

Our goal is to link the point clouds to energy performance certificate data[26]. The certificates are available for a subset of properties in England and Wales. There are currently around 25.60 million domestic EPCs and 1.11 non-domestic EPCs in the database[6]. The main label is an energy rating between A for most efficient and G for least efficient. However, the EPC database includes more detailed information, such as roof insulation efficiency or window efficiency, which are also added as energy attributes. The data is available under the OGL, except for the Ordnance Survey address data entries, which are therefore excluded from our dataset.

To link the EPC with the building point clouds, we use the Unique Property Reference Numbers (UPRN), which exist for most of the UK's buildings[27]. EPC data is structured by local authority districts (LAD). Accordingly, we also download LADs' boundary shape files provided by the Office for National Statistics[28] to limit the point cloud generation to geographic Areas of Interest (AOI).

Lastly, we aim at generating a dataset of representative buildings. We use the 2011 Rural Urban Classification[29]. The classification distinguishes ten rural urban classes ranging from "rural hamlets and isolated dwellings in a sparse setting" to "urban major conurbation". We assign one of these classes to each of the footprints. This allows us to select AOIs so that our dataset possesses the same rural urban distribution as entire England. The smallest available geographic entities are the Output Areas (OA). OAs entail an average resident population of approximately 300 people[33]. Therefore, we obtain the OA boundary shape files[30] as the last data source.

**Storing data in a database**

A key element of the dataset generation process is a postgres[34] database to store the obtained data. We use the postGIS[35] extension for geo-data and the pgpointcloud[36] extension for the point cloud data, as well as gdal[37] and pdal[38] to insert the data into the database, respectively. The program itself is written in Python3. The program manipulates data and interacts with the postgres database. The code and utilized Python libraries can be found at https://github.com/kdmayer/PointER. The framework was set up in a Singularity[39] container. Most of the data is inserted into the database before running the point cloud generation script. However, point cloud and EPC data are imported periodically during the runtime for one AOI only. This reduces the amount of data in the database and speeds up the cropping process.

**Selecting Areas of Interest**

To generate point clouds for a representative subset of England's building stock, we choose a geographically diverse set of LADs. We first select Coventry, Westminster, Oxford, and Peterborough, similarly to Mayer et al.[21]. Furthermore, we use the RUC for small geographies to calculate England's RUC distribution and all LAD's distribution. We choose LADs with the goal to reach a RUC distribution similar to England in our dataset. To account for the geographic diversity of buildings in England, we select LADs from across the country including coastal and interior regions. Figure 3 visualizes the geographic distribution of selected LADs. Furthermore, columns one and two of Table 2 give and overview of selected LADs, sorted by district code.

Our data includes other properties besides rural urban classification, such as building age, height, size and energy related features, which are relevant to achieve a representative dataset, too. While we do not take this data into account for the selection of regions to reduce complexity, we analyzed these properties in retrospect and included the result in the chapter Technical Validation.

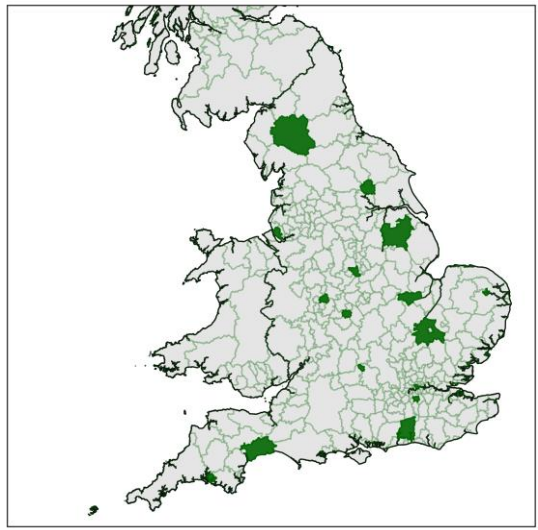

*Figure 3: Map of UK's Local Authority Districts (LADs) and the 16 LADs selected for our dataset highlighted in green*

**Generating building point clouds**

Building point clouds are generated for one AOI at a time. Figure 4 depicts the eight steps of the program. We used python to process the steps and to execute SQL queries in the postgres database, whereas steps one to five use postgres and steps five to seven run in python itself.

Step 1: The program starts by inserting the respective point cloud and EPC data into the database. This reduces the amount of data simultaneously in the database and speeds up GIS-functions such as the intersection function.

Step 2: The program runs batches of footprints iteratively to avoid memory issues. In our experience, a batch size of 500 works well, leading to 120 iterations for an AOI with 60.000 footprints.

Therefore, the program selects all footprints inside the AOI and saves them as a materialized view in the second step. This reduces the computationally expensive intersection of footprints in the AOI at each of the thousands of iterations.

Step 3: To account for small deviations in the spatial correlation between the footprint and the point cloud data, a buffer of 0.5 m is added to each footprint. In Figure 4, this buffer is indicated by the footprint's black outlines.

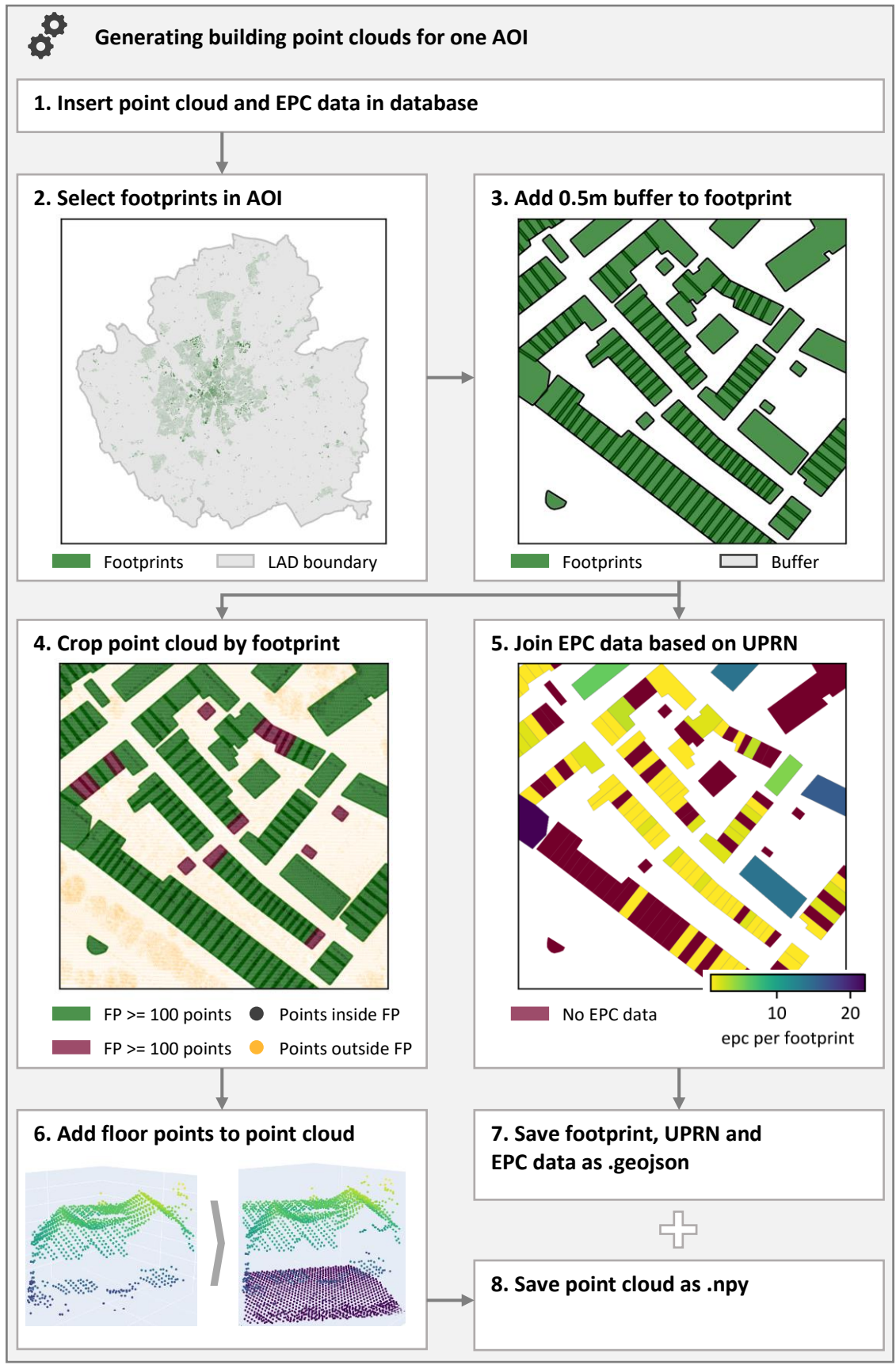

*Figure 4: Overview and visualization of the program's steps for generating building point clouds and their energy characteristics for one AOI*

Step 4: Next, all LiDAR points within footprints are selected and assigned to the respective footprint. In Figure 4 points intersecting with a footprint are plotted in black. Points outside footprints are plotted in orange. In step 4, we also apply a minimum threshold of 100 points to the footprints and filter out buildings with less points than this threshold. This is indicated by red polygons in Figure 4. We use this pre-selection to avoid point clouds that do not provide a useful geometric representation of a building and would reduce the quality of the dataset.

Step 5: In step 5, we match EPC data to the buildings by UPRN. Currently, only a subset of buildings holds EPCs which leads to a significant number of houses without an energy attributes (red polygon in Figure 4). Furthermore, it is possible that one footprint is allocated with multiple EPC, if a building contains multiple dwellings. In addition, some footprints do not have a linked UPRN. This fact is discussed in more detail in section Technical Validation.

Step 6: Airborne LiDAR data only covers the buildings' roofs and fragments of the walls. Therefore, we add floor points with in a raster of 0.5 m to the point cloud, because we expect that this improves the representation of the building. The step is conducted in python using the shapely library.

Step 7 and 8: In the end, the point clouds are saved in ".npy" format and the footprint polygons and EPC data are saved in a ".geojson" file.

## Data Records

### Dataset size and content

The PointER dataset comprises more than one million point clouds. The selected regions include almost 1.4 million buildings, however our point cloud threshold of 100 points per building point cloud eliminates of around 25% of buildings. As Table 2 shows, this effect is strongest in Plymouth, Coventry, Erewash, and Sutton.

Table 2: Overview of number of footprints with point cloud, UPRN and EPC data in the dataset

| LAD Code | Name | # footprints | # FP with point clouds | # FP with UPRN | # FP with UPRN & EPC | FP with full info |
|---|---|---|---|---|---|---|
| E06000014 | York | 85.551 | 89% | 84% | 47% | 44% |
| E06000026 | Plymouth | 154.151 | 62% | 60% | 38% | 33% |
| E06000031 | Peterborough | 80.532 | 80% | 84% | 56% | 46% |
| E07000012 | South Cambridgeshire | 69.179 | 83% | 67% | 41% | 38% |
| E07000030 | Eden | 35.264 | 88% | 58% | 32% | 29% |
| E07000036 | Erewash | 72.133 | 65% | 62% | 35% | 29% |
| E07000040 | East Devon | 76.142 | 83% | 68% | 41% | 36% |
| E07000142 | West Lindsey | 51.758 | 83% | 66% | 39% | 36% |
| E07000148 | Norwich | 66.050 | 75% | 73% | 47% | 42% |
| E07000178 | Oxford | 49.882 | 69% | 85% | 54% | 37% |
| E07000227 | Horsham | 63.711 | 83% | 68% | 39% | 36% |
| E08000012 | Liverpool | 202.656 | 76% | 88% | 53% | 42% |
| E08000026 | Coventry | 138.458 | 64% | 86% | 51% | 33% |
| E08000030 | Walsall | 116.036 | 81% | 84% | 47% | 41% |
| E09000029 | Sutton | 96.167 | 65% | 57% | 30% | 28% |
| E09000033 | Westminster | 27.371 | 96% | 84% | 54% | 54% |
| *Absolute sum* | | 1.385.041 | **1.040.425** | 1.038.406 | 619.554 | **518.992** |

Furthermore, only around 45% of footprints can be linked with EPC data by UPRN. We observe a great variance with 56% in Peterborough and 30% in Sutton. As a result, our final dataset contains around half a million point clouds with energy feature data.

**Dataset structure**

Table 3 gives a schematic overview of the dataset's folder structure. The dataset contains one result folder for each of the 16 select AOIs named according to the AOI's LAD code, e.g. E06000014 for York. Each result folder contains one sub-folder with all building point clouds in python numpy ".npy" format. Building point cloud files are named according to their footprint's centroid's coordinate in the spatial reference system EPSG 27700, i.e. "XCOORDINATE_YCOORDINATE.npy". Point coordinates of a point cloud are in EPSG 27700. In addition, there is a "final_result_AOI_CODE.json" file that maps the building point cloud file to the EPC data. Finally, a summary of the number of footprints with point cloud and EPC information of the AOI is stored in the "production_metrics_AOI_CODE.json" file.

*Table 3: Folder structure of published dataset*

| Result folder | Sub-folder | Point cloud files |
|---|---|---|
| E06000014 | npy_raw | 448435.3287834528_209292.45212838988.npy |
|  |  | 448446.4391428226_209267.82503584144.npy |
|  |  | … |
|  | final_result_E06000014.json |  |
|  | production_metrics_E06000014.json |  |
| E06000026 | … |  |
| … | … |  |
| E09000033 | … |  |

Table 4 visualizes the structure of the "final_result_.json" files. The first four columns originate from the point cloud generation process. They connect the footprint with the resulting point cloud file. The information about number of points refers to the building point cloud before adding floor points. This can be used to filter out footprints with too little or too many points depending on the use case. Furthermore, the UPRN column enables linking data to the point cloud.

*Table 4: Structure of feature table provided in final_result.json*

| Data type | Column name | Description |
|---|---|---|
| **Point cloud data** | id_fp | *Unique footprint identifier* |
|  | pc_file_name | *Filename of point cloud* |
|  | num_p_in_pc | *Number of points in point cloud* |
|  | **uprn** | *UPRN of Verisk dataset* |
| **Linked EPC data** | LMK_KEY | *Unique EPC identifier* |
|  | BUILDING_REFERENCE_NUMBER | *EPC building reference number* |
|  | CURRENT_ENERGY_RATING | *Energy rating and efficiency score* |
|  | POTENTIAL_ENERGY_RATING |  |
|  | CURRENT_ENERGY_EFFICIENCY |  |
|  | POTENTIAL_ENERGY_EFFICIENCY |  |
|  | PROPERTY_TYPE | *80 columns with more detailed building and energy information* |
|  | … |  |
|  | **UPRN** | *UPRN of EPC dataset* |
|  | UPRN_SOURCE |  |

In the provided dataset, we combine point clouds with EPC data. The EPC features consist of a unique identifier for each EPC entry, current and potential energy rating, current and potential energy efficiency as well as 80 more features with detailed building and energy information. Point cloud data and EPC are linked through the Verisk footprint UPRN and the EPC UPRN.

## Technical Validation

This section discusses the technical quality of the dataset in four aspects. First, information on the quality of the input data is summarized from the respective references. Second, we present an estimation of the resulting point cloud quality by conducting a manual assessment on a subset of 5000 point clouds. Next, we evaluate the linkage of point cloud features through UPRN. The last part presents a representativity study to verify that the chosen footprints in the dataset are representative for England.

### Quality of input data

The National LiDAR Programme covers the entire area of England. We download the LiDAR point cloud data with resolution of 1m which translates to an average point cloud density of 1 point per square meter. The National LiDAR Programme offers a metadata dashboard with details on each survey's mission dates and quality metrics[40]. The average ground truth error across 1215 surveys is 3.66 cm. Furthermore, 97.1% of LiDAR points from overlapping flightlines are less than 15 cm different in elevation[40]. Verisk's UKBuildings edition 13 online version dataset includes 28,733,631 buildings from the whole of Great Britain as well as the Belfast urban area. Verisk conducts an internal data quality analysis on accuracy and completeness, but only plans to publish the results in future releases. Therefore, we conducted a qualitative assessment by visualizing the footprint polygons and LiDAR data on Google aerial as depicted in Figure 2.

Information about the quality of EPC data can be found in the publication's technical notes[41]. The EPC dataset contains around 60% of the housing stock in England and just less than 60% in Wales[41]. The proportion is similar across all regions in England[41]. The energy assessment of individual buildings is conducted by energy assessors, who are responsible for the robustness of the data in relation to individual buildings. In addition, there are validation checks as the data is uploaded on the registry[41].

Finally, Local Authority District boundaries, Output Area boundaries and Rural Urban Classification affect the selection of buildings for point cloud generation, but not the building point clouds itself.

### Quality assessment of building point clouds

Our approach uses building outlines to crop the point clouds. Hence, errors can arise from spatial or temporal mismatch between point cloud data and building footprints. Therefore, we conduct a manual inspection of the resulting building point clouds. To this end, we randomly select a subset of 5000 point clouds assuming that the quality of the subset can be extrapolated to the entire dataset. First, we classify the point clouds into "suitable", "inspection required" and "unsuitable" based on a 2D visualization. In a second round, we inspect the 3D representation as well as an aerial image of the "inspection required" buildings and classify them into "suitable after inspection" and "unsuitable". Figure 5 gives examples for building point clouds of different quality.

Around 5% of the point clouds are of low-quality, meaning, that the building is not recognizable. There are a number of reasons for this. Some buildings miss points on the roof, usually caused by surrounding structures. Other buildings are covered by vegetation which results in a chaotic point structure. Most of these buildings are small buildings such as garages or auxiliary buildings. Furthermore, a poor spatial alignment between LiDAR data and building outline can lead to missing roof structures. The different temporal representations in the two datasets leads to some point clouds that appear to be vegetation or a construction site instead of a building. Finally, for some of the geographic areas, LiDAR points of two surveys overlap, which can result in distorted building point clouds in a few cases.

After reassessing "inspection required" point clouds in detail, 478 out of 650 buildings were classified as "suitable after inspection". The vast number of these point clouds can be divided into two cases. First, there are buildings that contain a small number of vegetation points covering the roof, but the overall building is clearly recognizable to the human eye. Second, the building point cloud appears slightly odd at first, but the building is part of narrow terrace house complex. Therefore, the these 478 point clouds can be described as edge cases. They make up around 10% of buildings. The remaining 85% of buildings are classified as suitable.

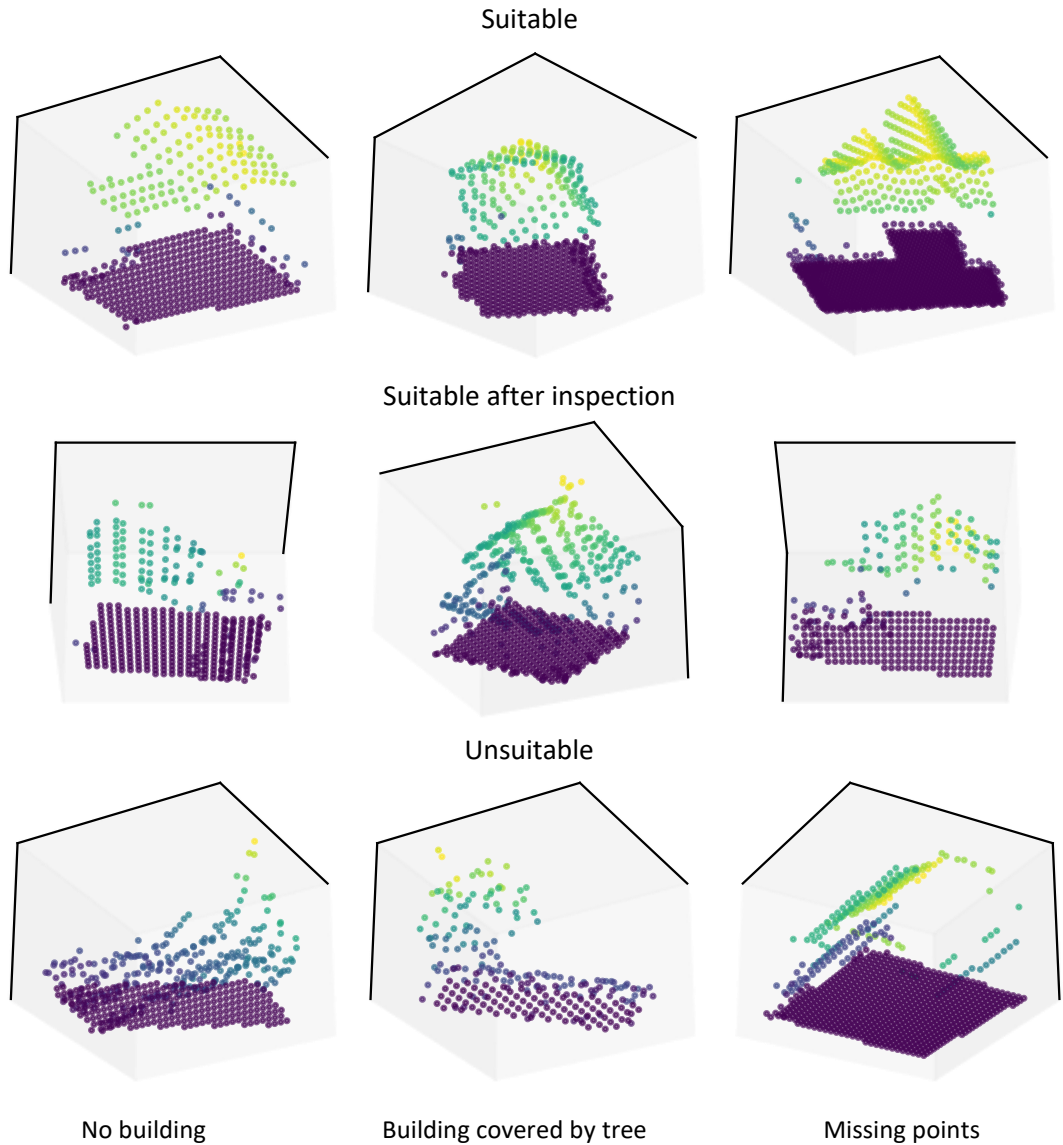

*Figure 5: Visualization of building point cloud examples of three assessment categories. Around 85% of building point clouds are "suitable", 10% are "suitable after inspection" and 5% are "unsuitable"*

**Point per Footprint**

Part of our approach is filtering out point clouds that contain less points than a defined threshold of 100 points. This is because buildings with fewer than 100 points are found to lack a rich representation. Figure 6 provides an impression of the number of LiDAR points per footprint for Westminster and Coventry. In Westminster (left) almost all buildings have more points than the threshold, except for a few garage or auxiliary garden buildings, which are visualized in red. On the other hand, the selected district in Coventry includes many buildings with less than 100 points. These are small terraced buildings that simply have a small footprint area and consequently do not contain enough points. Some of the footprints (yellow) comprise just enough points to be above the threshold. Furthermore, garages and auxiliary buildings, too are mostly below 100 points. While the point cloud threshold leads to a higher number of well-represented buildings, it also leads to a bias by removing small buildings. This can also be observed in the representativity analysis illustrated in Figure 8.

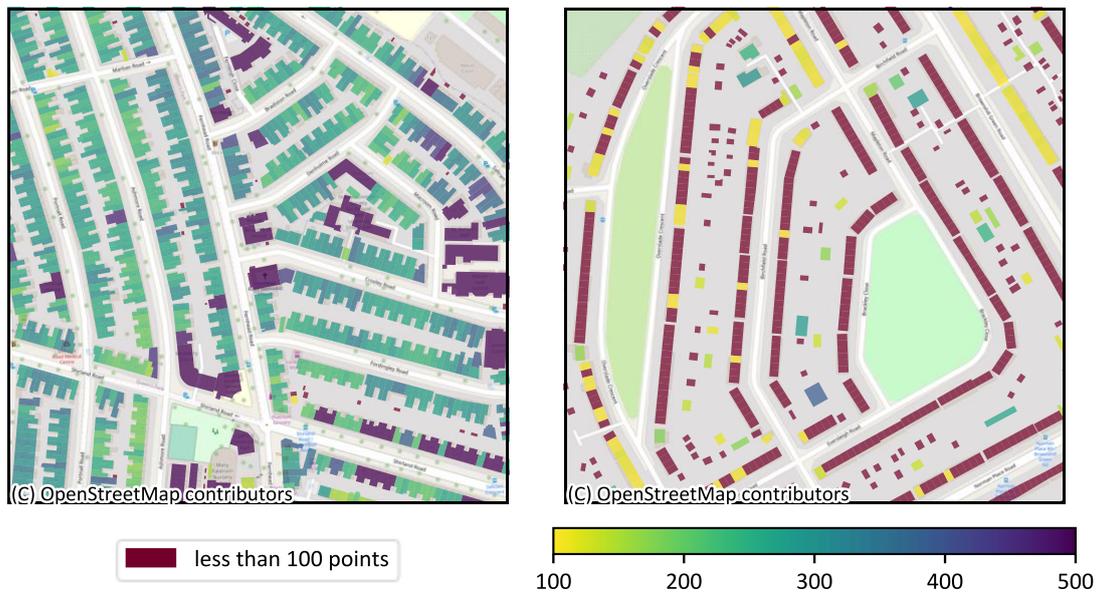

*Figure 6: Building footprints of Westminster (left) and Coventry (right) with number of points within their boundary. Westminster has only few buildings with less than 100 points (visualized in red), whereas Coventry displays larger numbers of those footprints in some areas of the city.*

**UPRN and EPC per Footprint**

To join the point cloud with EPC data we use the footprint dataset's UPRN feature. As an alternative, UPRN can be allocated to footprints through spatial intersection. However, the spatial intersection approach leads to a lower number of buildings with UPRN. Therefore, we decide to use the UPRN feature provided by Verisk. Figure 7 shows the number of UPRN and EPC entries per footprint for buildings in our dataset.

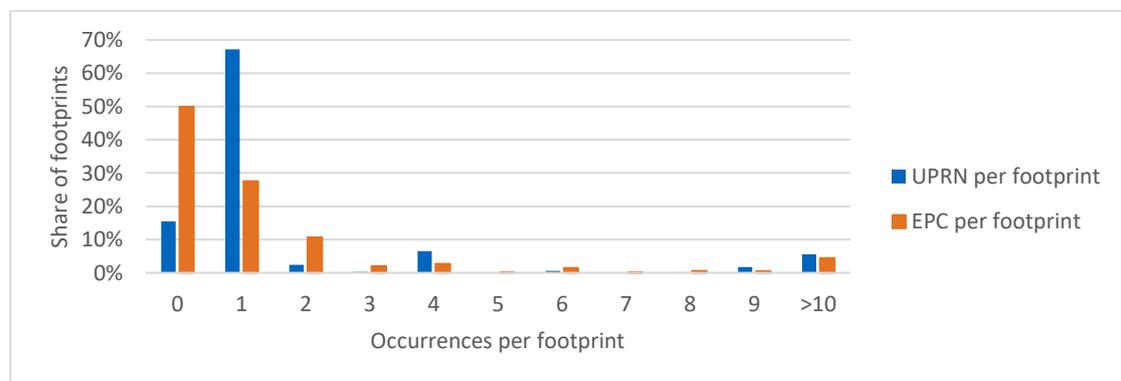

*Figure 7: Number of UPRN and EPC data points per footprint for our dataset.*

In our PointER dataset 1,040,425 or 84.54% of the footprints have at least one UPRN. Furthermore, some footprints are linked to more than one UPRN. As the UPRN refers to a dwelling not to a building, this can be the case for multi-dwelling units. Furthermore, multiple UPRN per footprint could also arise due to erroneous UPRN allocation in some cases. Overall, we link 27.14% of footprints with exactly one EPC entry and 56.91% of footprints remain without EPC data. Moreover, 15.95% of footprints have multiple EPC values. In our dataset, in the final result data frame, this is reflected by multiple rows that have identical point cloud filenames, but different EPC data. Multiple EPC links require an EPC selection approach. For example, the selection of the average or lowest EPC rating linked to a footprint could be used.

**Representativity of point clouds**

This section evaluates our subset's representativity in terms of RUC, area, height, age class and EPC rating. The distribution in our dataset in comparison to England is visualized in Figure 8.

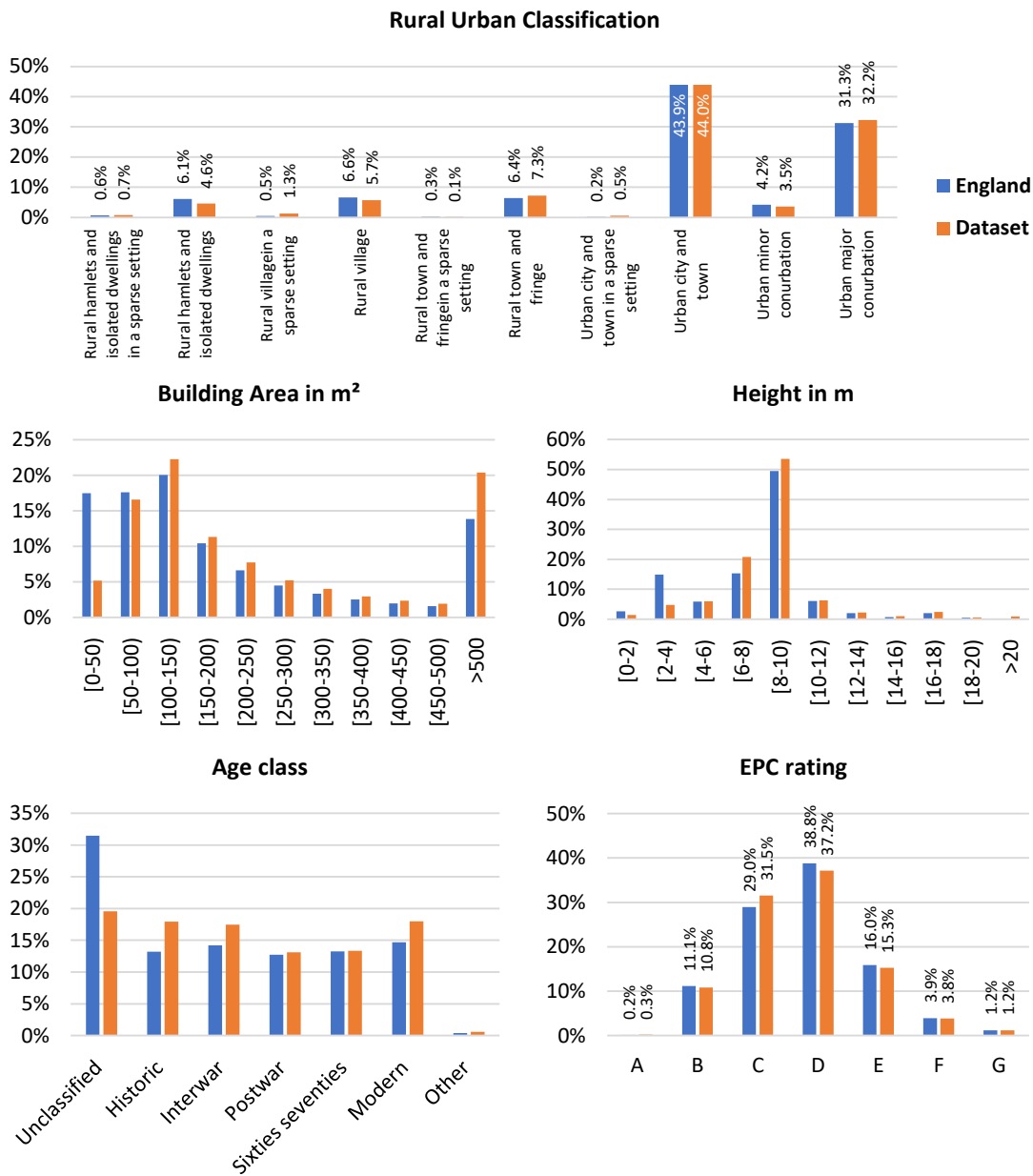

*Figure 8: Distribution of Rural Urban Classification, building area, height, age class and EPC rating of our dataset in comparison to all buildings in England included in UK Buildings dataset*

The graph at the top shows the asymmetry between the 10 rural-urban classes, which leads to the challenge of achieving exact alignment of the RUC distribution. Nevertheless, the overall representation is similar, especially for the largest classes "Urban city and town", "Urban major conurbation", "Rural town and fringe", "Rural village" and "Rural hamlets and isolated dwellings". The building characteristics area and height are shown in the two center graphs and display high agreement, except for small buildings with areas under 50m² and a height of 2-4m. This can be attributed to our approach of filtering out buildings that contain too few points. Furthermore, the figure depicts a difference in the age class distribution, where our dataset comprises less unclassified buildings than the England dataset. Instead, our dataset contains slightly more modern, interwar, and historic buildings. Finally, a representative distribution of EPC ratings is key for the application in building energy modeling. The bar graph on the bottom right in Figure 8 indicates a similar distribution of EPC ratings of our dataset in comparison to all of England's footprints. Therefore, we conclude, that by our approach of selecting building footprints based on RUC, we derive a subset of buildings that is representative of England's building stock.

## Usage Notes

The PointER dataset can be downloaded from https://doi.org/10.14459/2023mp1713501.

As mentioned in the paragraph on UPRN and EPC per footprint, some footprints are linked to multiple energy characteristics, because a building can contain multiple dwellings. When users require linking point clouds to unique energy features, they first need to apply a selection process. As there are multiple approaches with advantages and disadvantages depending on the use case, we leave this step to future users.

Depending on the application, the dataset might need to be normalized (e.g. in a point cloud deep learning pipeline). Currently, coordinates are in the metric coordinate reference system EPSG 27700, the Ordnance Survey National Grid reference system. Buildings can be normalized in relation to the largest building if scale preservation is required. However, most point clouds will consequently only occupy a fraction of the normalized space, because large buildings are less common, as visible in the height and area distribution displayed in Figure 8.

All data is licensed under the Open Government License, except for the building footprints. Hence, users can exploit the dataset both commercially and non-commercially, but have to acknowledge the sources of the data. The UKBuildings footprint data is provided by Verisk under a private license. Therefore, we can use building footprints to generate point clouds, but we cannot include them in this dataset for download. To extend the dataset to other regions, we recommend contacting Verisk for their UKBuildings dataset. Alternatively, other footprint data, such as OSM can be used. When using OSM, regions with high data quality and coverage should be selected.

This dataset can be used in a range of applications. For example, explorative studies can investigate the explanatory power of LiDAR data for building energy characteristics, e.g. through the application of deep learning methods. Recent advances in deep learning methods for point clouds and their application are promising[42–45] and we expect building point clouds to include significant information. Hence, studies can build point cloud classification models and predict EPC labels for all buildings in England and the UK.

Using our open source code, building point clouds can be generated for any location with available building footprint and LiDAR data. This way, future studies can be conducted across multiple countries. In addition, our dataset could be coupled with additional data sources such as aerial images, street view images, historic energy usage, or socio-demographic data.

Although our dataset is motivated by the challenge of modeling building energy efficiency, it is also relevant for applications outside of this field. Building point clouds can be used to evaluate architectural features or to support urban planning activities. Through the standardized Unique Property Reference Number (UPRN)[27] other datasets can easily be linked to the building point clouds.

## Code Availability

The code used for generating building point clouds is available at https://github.com/kdmayer/PointER. The repository includes a detailed description of software and python packages used, as well as their versions.

## Acknowledgements

This research was supported by Stanford's Bits&Watts initiative in collaboration with E.ON Innovation. This work was also supported by a fellowship of the German Academic Exchange Service (DAAD).

## Author contributions

Sebastian Krapf: Methodology, Software, Validation, Formal analysis, Investigation, Resources, Data Curation, Writing - Original Draft, Writing - Review & Editing, Visualization, Funding acquisition;

Kevin Mayer: Conceptualization, Methodology, Software, Validation, Resources, Data Curation, Writing - Review & Editing, Supervision, Project administration;

Martin Fischer: Writing - Review & Editing, Supervision, Funding acquisition

## Competing interests

The authors declare no competing interests.